\documentclass[11pt,a4paper]{article}
\usepackage[hyperref]{emnlp-ijcnlp-2019}
\usepackage{times}
\usepackage{latexsym}
\usepackage{amsmath}
\usepackage{amssymb}
\usepackage{graphicx}
\usepackage{bbm}

\usepackage{url}
\usepackage{footnote}
\makesavenoteenv{tabular}
\makesavenoteenv{table}

\aclfinalcopy 

\title{Patient Knowledge Distillation for BERT Model Compression}

\author{\textbf{Siqi Sun},\hspace{1mm} \textbf{Yu Cheng},\hspace{1mm} \textbf{Zhe Gan},\hspace{1mm} \textbf{Jingjing Liu}\\
Microsoft Dynamics 365 AI Research\\
{\tt \small{ \{Siqi.Sun,Yu.Cheng,Zhe.Gan,jingjl\}@microsoft.com}}}

\date{}

\begin{document}
\maketitle
\begin{abstract}
Pre-trained language models such as BERT have proven to be highly effective for natural language processing (NLP) tasks. However, the high demand for computing resources in training such models hinders their application in practice.
In order to alleviate this resource hunger in large-scale model training, we propose a Patient Knowledge Distillation approach to compress an original large model (teacher) into an equally-effective lightweight shallow network (student). Different from previous knowledge distillation methods, which only use the output from the last layer of the teacher network for distillation, our student model \textit{patiently} learns from multiple intermediate layers of the teacher model for incremental knowledge extraction, following two strategies: ($i$) PKD-Last: learning from the last $k$ layers; and ($ii$) PKD-Skip: learning from every $k$ layers. These two patient distillation schemes enable the exploitation of rich information in the teacher's hidden layers, and encourage the student model to \textit{patiently} learn from and imitate the teacher through a multi-layer distillation process. 
Empirically, this translates into improved results on multiple NLP tasks with significant gain in training efficiency, without sacrificing model accuracy.\footnote{Code will be avialable at \url{https://github.com/intersun/PKD-for-BERT-Model-Compression}.}
\end{abstract}


\section{Introduction}
Language model pre-training has proven to be highly effective in learning universal language representations from large-scale unlabeled data.  ELMo~\cite{peters2018deep}, GPT~\cite{radford2018improving} and BERT~\cite{devlin2018bert} have achieved great success in many NLP tasks, such as sentiment classification~\cite{socher2013recursive}, natural language inference~\cite{williams2017broad}, and question answering~\cite{lai2017race}. 

Despite its empirical success, BERT's computational efficiency is a widely recognized issue because of its large number of parameters. For example, the original BERT-Base model has 12 layers and 110 million parameters. Training from scratch typically takes four days on 4 to 16 Cloud TPUs. Even fine-tuning the pre-trained model with task-specific dataset may take several hours to finish one epoch.
Thus, reducing computational costs for such models is crucial for their application in practice, where computational resources are limited. 

Motivated by this, we investigate the redundancy issue of learned parameters in large-scale pre-trained models, and propose a new model compression approach, \emph{Patient Knowledge Distillation} (Patient-KD), to compress original teacher (e.g., BERT) into a lightweight student model without performance sacrifice. In our approach, the teacher model outputs probability logits and predicts labels for the training samples (extendable to additional unannotated samples), and the student model learns from the teacher network to mimic the teacher's prediction. 

Different from previous knowledge distillation methods~\cite{hinton2015distilling, sau2016deep, lu2017knowledge}, we adopt a \emph{patient} learning mechanism: instead of learning parameters from only the last layer of the teacher, we encourage the student model to extract knowledge also from previous layers of the teacher network. We call this `Patient Knowledge Distillation'. This patient learner has the advantage of distilling rich information through the deep structure of the teacher network for multi-layer knowledge distillation. 

We also propose two different strategies for the distillation process: ($i$) PKD-Last: the student learns from the last $k$ layers of the teacher, under the assumption that the top layers of the original network contain the most informative knowledge to teach the student; and ($ii$) PKD-Skip: the student learns from every $k$ layers of the teacher, suggesting that the lower layers of the teacher network also contain important information and should be passed along for incremental distillation.  

We evaluate the proposed approach on several NLP tasks, including Sentiment Classification, Paraphrase Similarity Matching, Natural Language Inference, and Machine Reading Comprehension. Experiments on seven datasets across these four tasks demonstrate that the proposed Patient-KD approach
achieves superior performance and better generalization than standard knowledge distillation methods~\cite{hinton2015distilling}, with significant gain in training efficiency and storage reduction while maintaining comparable model accuracy to original large models. To the authors' best knowledge, this is the first known effort for BERT model compression. 




\section{Related Work}
\paragraph{Language Model Pre-training} 
Pre-training has been widely applied to universal language representation learning. Previous work can be divided into two main categories: ($i$) feature-based approach; ($ii$) fine-tuning approach. 

Feature-based methods mainly focus on learning: 
($i$) context-independent word representation (e.g., word2vec~\cite{mikolov2013distributed}, GloVe~\cite{pennington2014glove}, FastText~\cite{bojanowski2017enriching}); ($ii$) sentence-level representation (e.g.,~\citet{kiros2015skip,conneau2017supervised,logeswaran2018efficient}); and ($iii$) contextualized word representation (e.g., Cove~\cite{mccann2017learned}, ELMo~\cite{peters2018deep}). Specifically, ELMo~\cite{peters2018deep} learns high-quality, deep contextualized word representation using bidirectional language model, which can be directly plugged into standard NLU models for performance boosting. 


On the other hand, fine-tuning approaches mainly pre-train a language model (e.g., GPT~\cite{radford2018improving}, BERT~\cite{devlin2018bert}) on a large corpus with an unsupervised objective, and then fine-tune the model with in-domain labeled data for downstream applications~\cite{dai2015semi,howard2018universal}. Specifically, BERT is a large-scale language model consisting of multiple layers of Transformer blocks~\cite{vaswani2017attention}. BERT-Base has 12 layers of Transformer and 110 million parameters, while BERT-Large has 24 layers of Transformer and 330 million parameters. 
By pre-training via masked language modeling and next sentence prediction, BERT has achieved state-of-the-art performance on a wide-range of NLU tasks, such as the GLUE benchmark~\cite{wang2018glue} and SQuAD~\cite{rajpurkar2016squad}.

However, these modern pre-trained language models contain millions of parameters, which hinders their application in practice where computational resource is limited. In this paper, we aim at addressing this critical and challenging problem, taking BERT as an example, \textit{i.e.,} how to compress a large BERT model into a shallower one without sacrificing performance. Besides, the proposed approach can also be applied to other large-scale pre-trained language models, such as recently proposed XLNet~\cite{yang2019xlnet} and RoBERTa~\cite{DBLP:journals/corr/abs-1907-11692}. 


\paragraph{Model Compression \& Knowledge Distillation}
Our focus is model compression, \textit{i.e.,} making deep neural networks more compact \cite{han2015deep_compression,Cheng2015circulant}. A similar line of work has focused on accelerating deep network inference at test time \cite{46320} and reducing model training time \cite{huang2016stochastic}. 

A conventional understanding is that a large number of connections (weights) is necessary for training deep networks \cite{Denil2013,dcnn}. However, once the network has been trained, there will be a high degree of parameter redundancy. Network pruning \cite{Han2015pruning,He_2017_ICCV}, in which
network connections are reduced or sparsified, is one common strategy for model compression. Another direction is weight quantization \cite{GongLYB14,2018arXiv180205668P}, in which connection weights are constrained to a set of discrete values, allowing weights to be represented by fewer bits. However, most of these pruning and quantization approaches perform on convolutional networks. Only a few work are designed for rich structural information such as deep language models \cite{prunetransformer}.

Knowledge distillation \cite{hinton2015distilling} aims to
compress a network with a large set of parameters into a compact and fast-to-execute model.
This can be achieved by training a compact model to imitate the soft output of a larger model. \citet{Romero15-iclr} further demonstrated that intermediate representations learned by the large model can serve as hints to improve the training process and the final performance of the compact model. \citet{Chen2016Net2NetAL} introduced techniques for efficiently transferring knowledge from an existing network to a deeper or wider network. More recently,~\citet{DBLP:journals/corr/abs-1904-09482} used knowledge from ensemble models to improve single model performance on NLU tasks.
\citet{tan2019multilingual} tried knowledge distillation for multilingual translation.
Different from the above efforts, 
we investigate the problem of compressing large-scale language models, and propose a novel \emph{patient knowledge distillation} approach to effectively transferring knowledge from a teacher to a student model.  

\section{Patient Knowledge Distillation}
In this section, we first introduce a vanilla knowledge distillation method for BERT compression (Section~\ref{sec:kd}), then present the proposed Patient Knowledge Distillation (Section~\ref{sec:patient_kd}) in details.

\paragraph{Problem Definition}
The original large teacher network is represented by a function $f(\mathbf{x};\mathbf{\theta})$, where $\mathbf{x}$ is the input to the network, and $\mathbf{\theta}$ denotes the model parameters. The goal of knowledge distillation is to learn a new set of parameters $\mathbf{\theta}^{\prime}$ for a shallower student network $g(\mathbf{x};\mathbf{\theta}^{\prime})$, such that the student network achieves similar performance to the teacher, with much lower computational cost. Our strategy is to force the student model to imitate outputs from the teacher model on the training dataset with a defined objective $L_{KD}$.  

\subsection{Distillation Objective} \label{sec:kd}
In our setting, the teacher $f(\mathbf{x};\mathbf{\theta})$ is defined as a deep bidirectional encoder, e.g., BERT, and the student $g(\mathbf{x};\mathbf{\theta}^{\prime})$ is a lightweight model with fewer layers. For simplicity, we use BERT$_k$ to denote a model with $k$ layers of Transformers. Following the original BERT paper~\cite{devlin2018bert}, we also use BERT-Base and BERT-Large to denote BERT$_{12}$ and BERT$_{24}$, respectively.

Assume $\{\mathbf{x}_i, \mathbf{y}_i\}_{i=1}^N$ are $N$ training samples, where $\mathbf{x}_i$ is the $i$-th input instance for BERT, and $\mathbf{y}_i$ is the corresponding ground-truth label. BERT first computes a contextualized embedding $\mathbf{h}_i = \text{BERT} (\mathbf{x}_i) \in \mathbb{R}^d$. Then, a softmax layer $\hat{\mathbf{y}}_i = P(\mathbf{y}_i | \mathbf{x}_i) = softmax(\mathbf{W} \mathbf{h}_i)$ for classification is applied to the embedding of BERT output, where $\mathbf{W}$ is a weight matrix to be learned. 

To apply knowledge distillation, first we need to train a teacher network. For example, to train a 12-layer BERT-Base as the teacher model, the learned parameters are denoted as:
\begin{align}
\hat{\theta}^{t}= \arg \min_\theta \sum_{i\in [N]} L_{CE}^t(\mathbf{x}_i,\mathbf{y}_i;[\theta_{\text{BERT}_{12}}, \mathbf{W}]) \,
\end{align}
where the superscript $t$ denotes parameters in the teacher model, $[N]$ denotes set $\{1, 2, \dots, N\}$, $L_{CE}^t$ denotes the cross-entropy loss for the teacher training, and $\theta_{\text{BERT}_{12}}$ denotes parameters of BERT$_{12}$. 

The output probability for any given input $\mathbf{x}_i$ can be formulated as: 
\begin{align}
    \hat{\mathbf{y}}_i &= {P}^t(\mathbf{y}_i|\mathbf{x}_i) = softmax\Big(\frac{\mathbf{W} \mathbf{h}_i}{T}\Big) \nonumber \\
    &= softmax\Big(\frac{\mathbf{W}\cdot \text{BERT}_{12}(\mathbf{x}_i; \hat{\theta}^{t})}{T}\Big)\, 
\end{align}
where ${P}^t(\cdot|\cdot)$ denotes the probability output from the teacher. $\hat{\mathbf{y}}_i$ is fixed as soft labels, and $T$ is the temperature used in KD, which controls how much to rely on the teacher's soft predictions. A higher temperature produces a more diverse probability distribution over classes~\cite{hinton2015distilling}. Similarly, let $\theta^s$ denote parameters to be learned for the student model, and ${P}^s(\cdot|\cdot)$ denote the corresponding probability output from the student model. Thus,
the distance between the teacher's prediction and the student's prediction can be defined as: 
\begin{align}
    L_{DS}= -\sum_{i\in [N]} & \sum_{c\in C} \Big[ P^t(\mathbf{y}_i = c|\mathbf{x}_i; \hat{\theta}^{t}) \cdot \nonumber \\
    & \log P^s(\mathbf{y}_i = c | \mathbf{x}_i; \theta^{s}) \Big] 
\end{align}
where $c$ is a class label and $C$ denotes the set of class labels.

Besides encouraging the student model to imitate the teacher's behavior, we can also fine-tune the student model on target tasks, where task-specific cross-entropy loss is included for model training: 
\begin{align}
    L_{CE}^s = - \sum_{i\in [N]} & \sum_{c\in C} \Big[ \mathbbm{1}[\mathbf{y}_i=c] \cdot \nonumber \\
    & \log P^s(\mathbf{y}_i =c| \mathbf{x}_i; \theta^s)\Big]  
\end{align}
Thus, the final objective function for knowledge distillation can be formulated as:
\begin{align}
\label{obj_kd}
L_{KD} = (1-\alpha) L_{CE}^s + \alpha L_{DS}\,
\end{align}
where $\alpha$ is the hyper-parameter that balances the importance of the cross-entropy loss and the distillation loss.

\begin{figure}[t!]
\centering
\includegraphics[width=\linewidth]{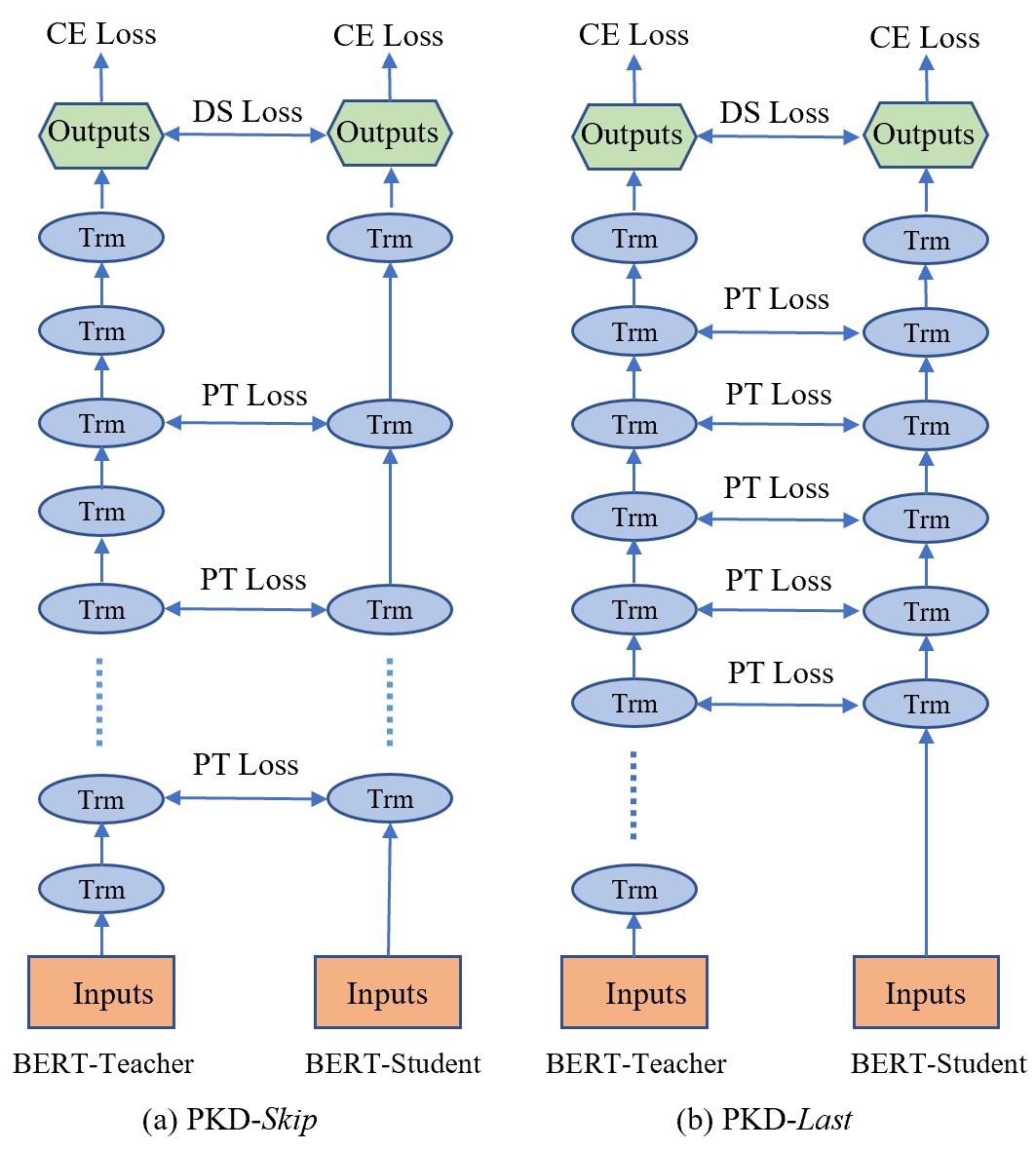}
\caption{Model architecture of the proposed Patient Knowledge Distillation approach to BERT model compression. (Left) PKD-Skip: the student network learns the teacher's outputs in every 2 layers. (Right) PKD-Last: the student learns the teacher's outputs from the last 6 layers. Trm: Transformer.}
\label{fig:model_arch}
\end{figure}

\subsection{Patient Teacher for Model Compression} \label{sec:patient_kd}
\begin{figure*}[t!]
\centering
\includegraphics[width=\linewidth]{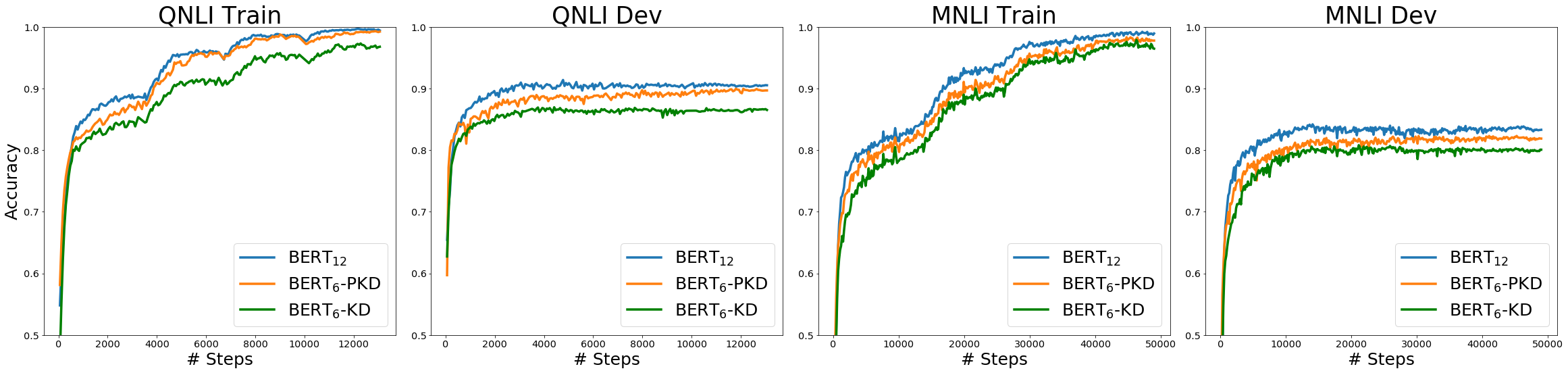}
\caption{Accuracy on the training and dev sets of QNLI and MNLI datasets, by directly applying vanilla knowledge distillation (KD) and the proposed Patient-KD-Skip. The teacher and the student networks are BERT$_{12}$ and BERT$_6$, respectively. The student network learned with vanilla KD quickly saturates on the dev set, while the proposed Patient-KD starts to plateau only in a later stage.  }
\label{fig:acc_dis}
\end{figure*}

Using a weighted combination of ground-truth labels and soft
predictions from the last layer of the teacher network, the student network can achieve comparable performance to the teacher model on the training set. 
However, with the number of epochs increasing, the student model learned with this vanilla KD framework quickly reaches saturation on the test set (see Figure~\ref{fig:acc_dis} in Section~\ref{sec:experiments}). 

One hypothesis is that overfitting during knowledge distillation may lead to poor generalization. To mitigate this issue, instead of forcing the student to learn only from the logits of the last layer, we propose a ``\emph{patient}'' teacher-student mechanism to distill knowledge from the teacher's intermediate layers as well. Specifically, we investigate two patient distillation strategies: ($i$) PKD-Skip: the student learns from every $k$ layers of the teacher (Figure~\ref{fig:model_arch}: Left); and ($ii$) PKD-Last: the student learns from the \emph{last} $k$ layers of the teacher (Figure~\ref{fig:model_arch}: Right).  

Learning from the hidden states of all the tokens is computationally expensive, and may introduce noise. In the original BERT implementation \cite{devlin2018bert}, prediction is performed by only using the output from the last layer's \texttt{[CLS]} token. In some variants of BERT, like SDNet ~\cite{zhu2018sdnet}, a weighted average of all layers' \texttt{[CLS]} embeddings is applied. In general, the final logit can be computed based on $\mathbf{h}_{\text{final}} = \sum_{j \in [k]} w_j \mathbf{h}_j$, where $w_j$ could be either learned parameters or a pre-defined hyper-parameter, $\mathbf{h}_j$ is the embedding of \texttt{[CLS]} from the hidden layer $j$, and $k$ is the number of hidden layers. Derived from this, if the compressed model can learn from the representation of \texttt{[CLS]} in the teacher's intermediate layers for any given input, it has the potential of gaining a generalization ability similar to the teacher model. 

Motivated by this, in our Patient-KD framework, the student is cultivated to imitate the representations only for the \texttt{[CLS]} token in the intermediate layers, following the intuition aforementioned that the \texttt{[CLS]} token is important in predicting the final labels. 
For an input $\mathbf{x}_i$, the outputs of the \texttt{[CLS]} tokens for all the layers are denoted as:
\begin{align}
    \mathbf{h}_i = [\mathbf{h}_{i,1}, \mathbf{h}_{i,2}, \dots, \mathbf{h}_{i,k}] = \text{BERT}_k(\mathbf{x}_i) \in \mathbb{R}^{k \times d}
\end{align}

We denote the set of intermediate layers to distill knowledge from as $I_{pt}$. Take distilling from BERT$_{12}$ to BERT$_6$ as an example. For the PKD-Skip strategy, $I_{pt} = \{2,4,6,8,10\}$; and  for the PKD-Last strategy, $I_{pt} = \{7,8,9,10,11\}$.
Note that $k=5$ for both cases, because the output from the last layer (e.g., Layer 12 for BERT-Base) is omitted since its hidden states are connected to the softmax layer, which is already included in the KD loss defined in Eqn. (\ref{obj_kd}). In general, for BERT student with $n$ layers, $k$ always equals to $n-1$.

The additional training loss introduced by the patient teacher is defined as the mean-square loss between the normalized hidden states:
\begin{align}
    \label{eqn:pt_loss}
    L_{PT} = \sum_{i=1}^N\sum_{j=1}^M \Big|\Big| \frac{\mathbf{h}_{i,j}^s}{||\mathbf{h}_{i,j}^s||_2} - \frac{\mathbf{h}^t_{i,I_{pt}(j)}}{||\mathbf{h}^t_{i,I_{pt}(j)}||_2}\Big|\Big|_2^2\,
\end{align}
where $M$ denotes the number of layers in the student network, $N$ is the number of training samples, and the superscripts $s$ and $t$ in $\mathbf{h}$ indicate the student and the teacher model, respectively. Combined with the KD loss introduced in Section~\ref{sec:kd}, the final objective function can be formulated as:
\begin{align}
    L_{PKD} = (1-\alpha) L_{CE}^s + \alpha L_{DS} + \beta L_{PT}\,
\end{align}
where $\beta$ is another hyper-parameter that weights the importance of the features for distillation in the intermediate layers. 

\section{Experiments} \label{sec:experiments}
In this section, we describe our experiments on applying the proposed Patient-KD approach to four different NLP tasks. Details on the datasets and experimental results are provided in the following sub-sections.
%
%
\begin{table*}[t!]
\centering
\begin{tabular}{|c c c c c c c c|} 
\hline
 Model & SST-2 & MRPC & QQP & MNLI-m & MNLI-mm & QNLI & RTE\\ 
 [0.5pt] & (67k) & (3.7k) & (364k) & (393k) & (393k) & (105k) & (2.5k)  \\
\hline\hline
 BERT$_{12}$ (Google) & 93.5 & 88.9/84.8 & 71.2/89.2 & 84.6 & 83.4 & 90.5 & 66.4 \\ 
 BERT$_{12}$ (Teacher) & 94.3 & 89.2/85.2 & 70.9/89.0 & 83.7 & 82.8 & 90.4 & 69.1\\
 \hline
 BERT$_6$-FT & 90.7	& 85.9/80.2	& 69.2/88.2	& 80.4	& 79.7	& 86.7	& 63.6 \\
 BERT$_6$-KD & 91.5 &	\textbf{86.2/80.6} &	70.1/88.8 &	80.2 &	79.8 &	88.3 &	64.7 \\
 BERT$_6$-PKD & \textbf{92.0} & 85.0/79.9 &	\textbf{70.7/88.9} &	\textbf{81.5} &	\textbf{81.0} &	\textbf{89.0} &	\textbf{65.5} \\  
 \hline
 BERT$_3$-FT &  86.4 &	80.5/\textbf{72.6} &	65.8/86.9 &	74.8 &	74.3 &	84.3 &	55.2\\
 BERT$_3$-KD & 86.9 &	79.5/71.1 &	67.3/87.6 &	75.4 &	74.8 &	84.0 &	56.2\\
 BERT$_3$-PKD & \textbf{87.5} & \textbf{80.7}/72.5 & \textbf{68.1/87.8} & \textbf{76.7} & \textbf{76.3} & \textbf{84.7} & \textbf{58.2} \\ 
 \hline
\end{tabular}
\caption{Results from the GLUE test server. The best results for 3-layer and 6-layer models are in-bold. Google's submission results are obtained from official GLUE leaderboard. BERT$_{12}$ (Teacher) is our own implementation of the BERT teacher model. FT represents direct fine-tuning on each dataset without using knowledge distillation. KD represents using a vanilla knowledge distillation method. And PKD represents our proposed Patient-KD-Skip approach. Results show that PKD-Skip outperforms the baselines on almost all the datasets except for MRPC. The numbers under each dataset indicate the corresponding number of training samples.}
\label{tab_glue}
\end{table*}

\subsection{Datasets}
We evaluate our proposed approach on Sentiment Classification, Paraphrase Similarity Matching, Natural Language Inference, 
and Machine Reading Comprehension tasks. 
For Sentiment Classification, we test on Stanford Sentiment Treebank (SST-2) ~\cite{socher2013recursive}. For Paraphrase Similarity Matching, we use Microsoft Research Paraphrase Corpus (MRPC) ~\cite{dolan2005automatically} and Quora Question Pairs (QQP)\footnote{https://data.quora.com/First-Quora-Dataset-Release-Question-Pairs} datasets.
For Natural Language Inference, we evaluate on Multi-Genre Natural Language Inference (MNLI) ~\cite{williams2017broad}, QNLI\footnote{The dataset is derived from Stanford Question Answer Dataset (SQuAD). }~\cite{rajpurkar2016squad}, and Recognizing Textual Entailment (RTE). 

More specifically, SST-2 is a movie review dataset with binary annotations, where the binary label indicates positive and negative reviews. MRPC contains pairs of sentences and corresponding labels, which indicate the semantic equivalence relationship between each pair. QQP is designed to predict whether a pair of questions is duplicate or not, provided by a popular online question-answering website Quora. MNLI is a multi-domain NLI task for predicting whether a given premise-hypothesis pair is entailment, contradiction or neural. Its test and development datasets are further divided into in-domain (MNLI-m) and cross-domain (MNLI-mm) splits to evaluate the generality of tested models. QNLI is a task for predicting whether a question-answer pair is entailment or not. Finally, RTE is based on a series of textual entailment challenges, created by General Language Understanding Evaluation (GLUE) benchmark~\cite{wang2018glue}.

For the Machine Reading Comprehension task, we evaluate on RACE~\cite{lai2017race}, a large-scale dataset collected from English exams, containing 25,137 passages and 87,866 questions. For each question, four candidate answers are provided, only one of which is correct. The dataset is further divided into RACE-M and RACE-H, containing exam questions for middle school and high school students. 

\subsection{Baselines and Training Details}
For experiments on the GLUE benchmark, since all the tasks can be considered as sentence (or sentence-pair) classification, we use the same architecture in the original BERT~\cite{devlin2018bert}, and fine-tune each task independently.

For experiments on RACE, we denote the input passage as $P$, the question as $q$, and the four answers as $a_1, \dots, a_4$. We first concatenate the tokens in $q$ and each $a_i$, and  arrange the input of BERT as \texttt{[CLS]} $P$ \texttt{[SEP]} $q+a_i$ \texttt{[SEP]} for each input pair $(P, q+a_i)$, where \texttt{[CLS]} and \texttt{[SEP]} are the special tokens used in the original BERT. In this way, we can obtain a single logit value for each $a_i$. At last, a softmax layer is placed on top of these four logits to obtain the normalized probability of each answer $a_i$ being correct, which is then used to compute the cross-entropy loss for modeling training. 

We fine-tune BERT-Base (denoted as BERT$_{12}$) as the teacher model to compute soft labels for each task independently, where the pretrained model weights are obtained from Google's official BERT's repo\footnote{https://github.com/google-research/bert}, and use 3 and 6 layers of Transformers as the student models (BERT$_{3}$ and BERT$_{6}$), respectively. We initialize BERT$_k$ with the first $k$ layers of parameters from pre-trained BERT-Base, where $k\in \{3, 6\}$. To validate the effectiveness of our proposed approach, we first conduct direct fine-tuning on each task without using any soft labels. In order to reduce the hyper-parameter search space, we fix the number of hidden units in the final softmax layer as 768, the batch size as 32, and the number of epochs as 4 for all the experiments, with a learning rate from \{5e-5, 2e-5, 1e-5\}. The model with the best validation accuracy is selected for each setting. 

Besides direct fine-tuning, we further implement a vanilla KD method on all the tasks by optimizing the objective function in Eqn. (\ref{obj_kd}). We set the temperature $T$ as \{5, 10, 20\}, $\alpha = \{ 0.2, 0.5, 0.7 \}$, and perform grid search over $T$, $\alpha$ and learning rate, to select the model with the best validation accuracy. For our proposed Patient-KD approach, we conduct additional search over $\beta$ from $\{10, 100, 500, 1000\}$ on all the tasks. Since there are so many hyper-parameters to learn for Patient KD, we fix $\alpha$ and $T$ to the values used in the model with the best performance from the vanilla KD experiments, and only search over $\beta$ and learning rate.

\begin{table*}[t!]
\centering
\begin{tabular}{|c c c c c c c c|} 
\hline
 Model & SST-2 & MRPC & QQP & MNLI-m & MNLI-mm & QNLI & RTE\\ [0.5ex] 
\hline\hline
BERT$_6$ (PKD-Last) & 91.9	& \textbf{85.1}/79.5	& 70.5/\textbf{88.9}	& 80.9	& \textbf{81.0}	& 88.2	& 65.0 \\
 BERT$_6$ (PKD-Skip) & \textbf{92.0} & 85.0/\textbf{79.9} &	\textbf{70.7/88.9} &	\textbf{81.5} &	\textbf{81.0} &	\textbf{89.0} &	\textbf{65.5} \\  
\hline
\end{tabular}
\caption{Performance comparison between PKD-Last and PKD-Skip on GLUE benchmark.}
\label{tab_glue_every_last}
\end{table*}

\subsection{Experimental Results}
%
%
We submitted our model predictions to the official GLUE evaluation server to obtain results on the test data. Results are summarized in Table~\ref{tab_glue}. Compared to direct fine-tuning and vanilla KD, our Patient-KD models with BERT$_3$ and BERT$_6$ students perform the best on almost all the tasks except MRPC. For MNLI-m and MNLI-mm, our 6-layer model improves 1.1\% and 1.3\% over fine-tune (FT) baselines; for QNLI and QQP, even though the gap between BERT$_6$-KD and BERT$_{12}$ teacher is relatively small, our approach still succeeded in improving over both FT and KD baselines and further closing the gap between the student and the teacher models.

Furthermore, in 5 tasks out of 7 (SST-2 (-2.3\% compared to BERT-Base teacher), QQP (-0.1\%), MNLI-m (-2.2\%), MNLI-mm (-1.8\%), and QNLI (-1.4\%)), the proposed 6-layer student coached by the patient teacher achieved similar performance to the original BERT-Base, demonstrating the effectiveness of our approach. Interestingly, all those 5 tasks have more than 60k training samples, which indicates that our method tends to perform better when there is a large amount of training data. 

For the QQP task, we can further reduce the model size to 3 layers, where BERT$_3$-PKD can still have a similar performance to the teacher model. The learning curves on the QNLI and MNLI datasets are provided in Figure~\ref{fig:acc_dis}. The student model learned with vanilla KD quickly saturated on the dev set, while the proposed Patient-KD keeps learning from the teacher and improving accuracy, only starting to plateau in a later stage. 

For the MRPC dataset, one hypothesis for the reason on vanilla KD outperforming our model is that the lack of enough training samples may lead to overfitting on the dev set. To further investigate, we repeat the experiments three times and compute the average accuracy on the dev set. 
We observe that fine-tuning and vanilla KD have a mean dev accuracy of 82.23\% and 82.84\%, respectively. Our proposed method has a higher mean dev accuracy of 83.46\%, hence indicating that our Patient-KD method slightly overfitted to the dev set of MRPC due to the small amount of training data. This can also be observed on the performance gap between teacher and student on RTE in Table \ref{tab:ablation-study}, which also has a small training set.

We further investigate the performance gain from two different patient teacher designs: PKD-Last vs. PKD-Skip. Results of both PKD variants on the GLUE benchmark (with BERT$_6$ as the student) are summarized in Table~\ref{tab_glue_every_last}. Although both strategies achieved improvement over the vanilla KD baseline (see Table~\ref{tab_glue}), PKD-Skip performs slightly better than PKD-Last. Presumably, this might be due to the fact that distilling information across every $k$ layers captures more diverse representations of richer semantics from low-level to high-level, while focusing on the last $k$ layers tends to capture relatively homogeneous semantic information.    


%
\begin{table}[t!]
\centering
\setlength\tabcolsep{2pt}
\begin{tabular}{|c c c c|} 
\hline
 Model & RACE & RACE-M & RACE-H \\
 \hline\hline
 BERT$_{12}$ \small{(Leaderboard)}  & 65.00 & 71.70 & 62.30 \\
 BERT$_{12}$ (Teacher) & 65.30 & 71.17 & 62.89 \\
  \hline
 BERT$_6$-FT & 54.32    & 61.07	& 51.54 \\
 BERT$_6$-KD & 58.74 &	64.69 &	56.29  \\
 BERT$_6$-PKD-Skip & \textbf{60.34} & \textbf{66.57} &	\textbf{57.78} \\  
 \hline
\end{tabular}
\caption{Results on RACE test set. BERT$_{12}$ (Leaderboard) denotes results extracted from the official leaderboard (\url{http://www.qizhexie.com//data/RACE_leaderboard}). BERT$_{12}$ (Teacher) is our own implementation. Results of BERT$_3$ are not included due to the large gap between the teacher and the BERT$_3$ student.}
\label{tab:race}
\end{table}
\begin{table*}[t!]
\small
\centering
\begin{tabular}{|c c c c c|} 
\hline
\# Layers & \# Param (Emb) & \# Params (Trm) & Total Params & Inference Time (s) \\ [0.5ex] 
\hline\hline
3 & 23.8M & 21.3M & 45.7M (2.40$\times$)  & 27.35 (3.73$\times$)\\
6 & 23.8M & 42.5M & 67.0M (1.64$\times$) & 52.51 (1.94$\times$)\\
12 & 23.8M & 85.1M & 109M (1$\times$) & 101.89 (1$\times$) \\
\hline
\end{tabular}
\caption{The number of parameters and inference time for BERT$_3$, BERT$_6$ and BERT$_{12}$. Parameters in Transformers (Trm) grow linearly with the increase of layers. Note that the summation of \# Param (Emb) and \# Param (Trm) does not exactly equal to Total Params, because there is another softmax layer with 0.6M parameters.}
\label{tab_inference}
\end{table*}

Results on RACE are reported in Table~\ref{tab:race}, which shows that the Vanilla KD method outperforms direct fine-tuning by 4.42\%, and our proposed patient teacher achieves further 1.6\% performance lift, which again demonstrates the effectiveness of Patient-KD.


\subsection{Analysis of Model Efficiency}
We have demonstrated that the proposed Patient-KD method can effectively compress BERT$_{12}$ into BERT$_6$ models without performance sacrifice. In this section, we further investigate the efficiency of Patient-KD on storage saving and inference-time speedup. Parameter statistics and inference time are summarized in Table \ref{tab_inference}. All the models share the same embedding layer with 24 millon parameters that map a 30k-word vocabulary to a 768-dimensional vector, which leads to 1.64 and 2.4 times of machine memory saving from BERT$_6$ and BERT$_3$, respectively.

To test the inference speed, we ran experiments on 105k samples from QNLI training set~\cite{rajpurkar2016squad}. Inference is performed on a single Titan RTX GPU with batch size set to 128, maximum sequence length set to 128, and FP16 activated. The inference time for the embedding layer is negligible compared to the Transformer layers. Results in Table \ref{tab_inference} show that the proposed Patient-KD approach achieves an almost linear speedup, 1.94 and 3.73 times for BERT$_6$ and BERT$_3$, respectively. 

\begin{table*}[t!]
\centering
\small
\begin{tabular}{|c c c c c c c c c c|} 
\hline
Setting & Teacher & Student &SST-2 & MRPC & QQP & MNLI-m & MNLI-mm & QNLI & RTE\\ [0.5ex] 
\hline\hline
N/A & N/A & BERT$_{12}$ (Teacher) & 94.3 & 89.2/85.2 & 70.9/89.0 & 83.7 & 82.8 & 90.4 & 69.1\\
N/A & N/A & BERT$_{24}$ (Teacher) & 94.3	& 88.2/84.3	& 71.9/89.4	& 85.7	& 84.8	& 92.2	& 72.8 \\  
\hline
\#1 & BERT$_{12}$ & BERT$_6$[Base]-KD  & 91.5 & 86.2/80.6 &	70.1/88.8 &	79.7 &	79.1 &	88.3 &	64.7 \\
\#2 & BERT$_{24}$ & BERT$_6$[Base]-KD  & 91.2	& 86.1/80.7	& 69.4/88.6	& 80.2	& 79.7	& 87.5	& 65.7\\
\#3 & BERT$_{24}$ & BERT$_6$[Large]-KD  & 89.6 & 79.0/70.0 & 65.0/86.7 & 75.3 & 74.6 & 83.4 & 53.7\\
\#4 & BERT$_{24}$ & BERT$_6$[Large]-PKD  & 89.8	& 77.8/68.3	& 67.1/87.9	& 77.2	& 76.7	& 83.8	& 53.2\\
\hline
\end{tabular}
\caption{Performance comparison with different teacher and student models. BERT$_6$[Base]/[Large] denotes a BERT$_6$ model with a BERT-Base/Large Transformer in each layer. For PKD, we use the PKD-Skip architecture.} 
\label{tab:ablation-study}
\end{table*}

\subsection{Does a Better Teacher Help?}
To evaluate the effectiveness of the teacher model in our Patient-KD framework, we conduct additional experiments to measure the difference between BERT-Base teacher and BERT-Large teacher for model compression. 

Each Transformer layer in BERT-Large has 12.6 million parameters, which is much larger than the Transformer layer used in BERT-Base. For a compressed BERT model with 6 layers, BERT$_6$ with BERT-Base Transformer (denoted as BERT$_6$[Base]) has only 67.0 million parameters, while BERT$_6$ with BERT-Large Transformer (denoted as BERT$_6$[Large]) has 108.4 million parameters. Since the size of the \texttt{[CLS]} token embedding is different between BERT-Large and BERT-Base, we cannot directly compute the patient teacher loss (\ref{eqn:pt_loss}) for BERT$_6$[Base] when BERT-Large is used as teacher. Hence, in the case where the teacher is BERT-Large and the student is BERT$_6$[Base], we only conduct experiments in the vanilla KD setting.

Results are summarized in Table~\ref{tab:ablation-study}.
When the teacher changes from BERT$_{12}$ to BERT$_{24}$ (\emph{i.e.}, Setting \#1 vs. \#2), there is not much difference between the students' performance. 
Specifically, BERT$_{12}$ teacher performs better on SST-2, QQP and QNLI, while BERT$_{24}$ performs better on MNLI-m, MNLI-mm and RTE. 
Presumably, distilling knowledge from a larger teacher requires a larger training dataset, thus better results are observed on  MNLI-m and MNLI-mm.

We also report results on using BERT-Large as the teacher and BERT$_6$[Large] as the student. Interestingly, when comparing Setting \#1 with \#3, BERT$_6$[Large] performs much worse than BERT$_6$[Base] even though a better teacher is used in the former case. The BERT$_6$[Large] student also has 1.6 times more parameters than BERT$_6$[Base]. One intuition behind this is that the compression ratio for the BERT$_6$[Large] model is 4:1 (24:6), which is larger than the ratio used for the BERT$_6$[Base] model (2:1 (12:6)). The higher compression ratio renders it more challenging for the student model to absorb important weights. 

When comparing Setting \# 2 and \#3, we observe that even when the same large teacher is used, BERT$_6$[Large] still performs worse than BERT$_6$[Base]. Presumably, 
this may be due to initialization mismatch. Ideally, we should pre-train BERT$_6$[Large] and BERT$_6$[Base] from scratch, and use the weights learned from the pre-training step for weight initialization in KD training. However, due to computational limits of training BERT$_6$ from scratch, we only initialize the student model with the first six layers of BERT$_{12}$ or BERT$_{24}$.
Therefore, the first six layers of BERT$_{24}$ may not be able to capture high-level features, leading to worse KD performance.

Finally, when comparing Setting \#3 vs. \#4, where for setting \#4 we use Patient-KD-Skip instead of vanilla KD, we observe a performance gain on almost all the tasks, which indicates Patient-KD is a generic approach independent of the selection of the teacher model (BERT$_{12}$ or BERT$_{24}$).

\section{Conclusion}
In this paper, we propose a novel approach to compressing a large BERT model into a shallow one via Patient Knowledge Distillation. To fully utilize the rich information in deep structure of the teacher network, our Patient-KD approach encourages the student model to patiently learn from the teacher through a multi-layer distillation process. Extensive experiments over four NLP tasks demonstrate the effectiveness of our proposed model. 

For future work, we plan to pre-train BERT from scratch to address the initialization mismatch issue, and potentially modify the proposed method such that it could also help during pre-training.
Designing more sophisticated distance metrics for loss functions is another exploration direction. We will also investigate Patient-KD in more complex settings such as multi-task learning and meta learning.


\bibliography{reference}
\bibliographystyle{acl_natbib}

\appendix


\end{document}